\documentclass[preprint]{elsarticle}

\usepackage[T1]{fontenc}
\usepackage[utf8]{inputenc}

\usepackage{cite}
\usepackage{booktabs, makecell}
\newcommand{\ra}[1]{\renewcommand{\arraystretch}{#1}}
\usepackage{multirow}
\usepackage{caption}
\usepackage{float}
\usepackage{tabularx}
\usepackage[flushleft]{threeparttable}
\usepackage{xcolor}
\usepackage{needspace}
\graphicspath{{./figures/}{./resources/}}
\DeclareGraphicsExtensions{.pdf,.png,.jpg,.jpeg}

\usepackage{wrapfig}
\usepackage{tikz}
\usetikzlibrary{shapes,arrows}
\usepackage{pgfplots}
\pgfplotsset{compat=1.9}
\usepackage{amssymb}
\usepackage{amsmath}
\usepackage{mathtools}

\usepackage{siunitx}
\usepackage{tabularx}
\usepackage{caption}
\usepackage{blindtext}
\usepackage{tikz}
\usetikzlibrary{shapes, arrows, decorations, positioning, calc, shadows, trees, mindmap}
\usepackage{listings}

\newcolumntype{L}[1]{>{\hsize=#1\hsize\raggedright\arraybackslash}X}%
\newcolumntype{R}[1]{>{\hsize=#1\hsize\raggedleft\arraybackslash}X}%
\newcolumntype{C}[1]{>{\hsize=#1\hsize\centering\arraybackslash}X}%
\newcolumntype{E}{>{\raggedleft\arraybackslash}X}%
\newcolumntype{M}{>{\centering\arraybackslash}X}%

\usepackage{xspace}
\makeatletter
\DeclareRobustCommand\onedot{\futurelet\@let@token\@onedot}
\def\@onedot{\ifx\@let@token.\else.\null\fi\xspace}

\def\eg{\emph{e.g}\onedot}

\def\ie{\emph{i.e}\onedot}

\makeatother

\usepackage{arydshln}
\usepackage{hyperref}

\usepackage{xcolor}

\definecolor{codegreen}{rgb}{0,0.6,0}
\definecolor{codegray}{rgb}{0.5,0.5,0.5}
\definecolor{codepurple}{rgb}{0.58,0,0.82}
\definecolor{backcolour}{rgb}{0.95,0.95,0.92}

\lstdefinestyle{mystyle}{
	backgroundcolor=\color{backcolour}, commentstyle=\color{codegreen},
	keywordstyle=\color{magenta},
	numberstyle=\tiny\color{codegray},
	stringstyle=\color{codepurple},
	basicstyle=\ttfamily\footnotesize,
	breakatwhitespace=false,         
	breaklines=true,                 
	captionpos=b,                    
	keepspaces=true,                 
	numbers=left,                    
	numbersep=5pt,                  
	showspaces=false,                
	showstringspaces=false,
	showtabs=false,                  
	tabsize=2
}

\usepackage{footnote}
\makesavenoteenv{tabular}
\makesavenoteenv{table}

\usepackage{pdflscape}

\journal{arXiv.org}

\begin{document}
	
	\begin{frontmatter}
		
		\title{Data Model Design and Feature Management for Emerging Energy ML Pipelines}
		
		\author[label1]{Gregor~Cerar\corref{cor1}}
		\ead{gregor.cerar@ijs.si}
		
		\author[label1]{Blaž~Bertalanič\corref{cor1}}
		\ead{blaz.bertalanic@ijs.si}
		
		\author[label1]{Anže~Pirnat}
		\ead{ap6928@student.uni-lj.si}
		
		\author[label1]{Andrej~Čampa}
		\ead{andrej.campa@ijs.si}
		
		\author[label1]{Carolina~Fortuna}
		\ead{carolina.fortuna@ijs.si}
		
		\affiliation[label1]{%
			organization={Department of Communication Systems, Jožef~Stefan~Institute},%
			addressline={Jamova~cesta~39},%
			city={Ljubljana},%
			country={Slovenia}%
		}
		
		\cortext[cor1]{These authors contributed equally}
		
		\begin{abstract}
			The digital transformation of the energy infrastructure enables new, data driven, applications often supported by machine learning models. However, domain specific data transformations, pre-processing and management in modern data driven pipelines is yet to be addressed.
			
			In this paper we perform a first time study on generic data models that are able to support designing feature management solutions that are the most important component in developing ML-based energy applications. We first propose a taxonomy for designing data models suitable for energy applications, explain how this model can support the design of features and their subsequent management by specialized feature stores. Using a short-term forecasting dataset, we show the benefits of designing richer data models and engineering the features on the performance of the resulting models. Finally, we benchmark three complementary feature management solutions, including an open-source feature store suitable for time series.
		\end{abstract}
		
		
	\end{frontmatter}
	
	%\linenumbers
	
	\section{Introduction}
	\label{sec:intro}
	
	With the transformation of the traditional power grid to the smart grid, the complexity of the system continues to evolve~\citep{fortuna2022}, especially with the penetration of smart meters (SM), energy management systems (EMS) and other intelligent electronic devices (IED) especially at the low voltage (LV) level of the grid. IEDs, together with EMS enable an innovative set of energy~\citep{dileep2020survey} and non-energy applications~\citep{chuang2020monitoring}. EMSes enable the  control of various assets in homes or buildings with limited knowledge of grid status. Example energy applications are energy cost optimization, matching consumption with self-production from renewable energy sources (RES), or by trying to help distribution system operator (DSO) or aggregator to reach their predictive performance curves.
	
	On the DSO side of the LV grid, the main challenges are represented by reliability and latency. In the case of controlling at substation level, the complete observability of the LV grid for that substation is of great importance. The data collected from all SMs in the grid of one substation would provide enough data to plan and minimize the possible congestion that may occur during the peak demand hours, or too high production of RESs that could lead to power quality issues (e.g. over-voltage).
	
	In the case of medium voltage (MV) and high voltage (HV) network wide-area measurement systems (WAMS) already monitor and collect data. However, the data is collected only to provide observability and to efficiently handle the critical situations that might result in catastrophic events, in the worst case, power outages. Since reliability is the most important factor, the penetration of auxiliary services in the MV and HV grid is low. However, the data collected in the LV grid could be processed and used to enrich the collected data at MV and HV levels.  The enriched data can be used to create a limited control loop that extends from the observability of the HV grid to control and make smaller adjustments as in the LV grid, all the way down to the prosumer.
	
	\begin{figure}[htbp]
		\centering
		\includegraphics[width=0.8\linewidth]{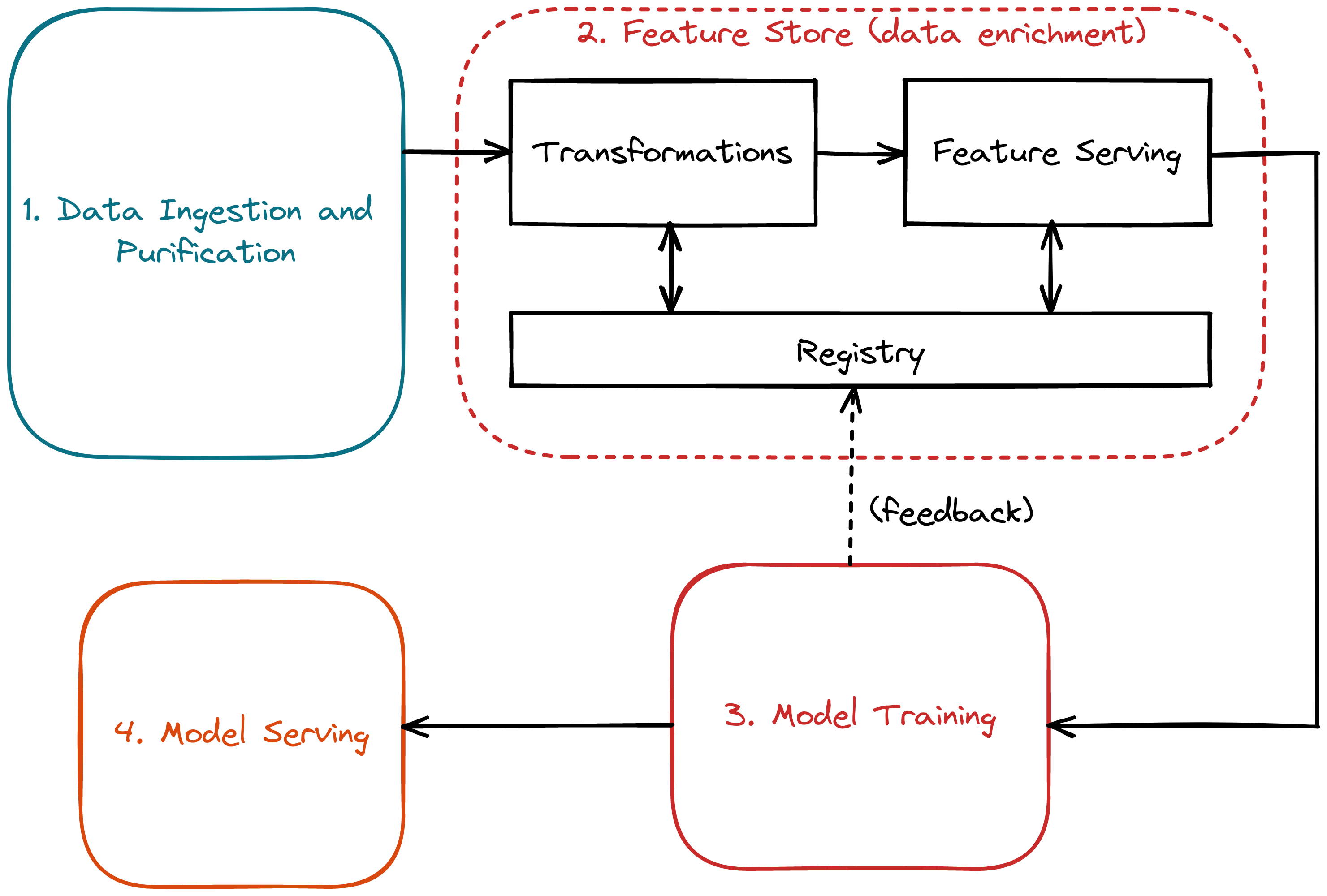}
		\caption{End-to-end infrastructure for machine learning model development and management.}
		\label{fig:data-infrastructure}
	\end{figure}
	
	While IEDs, WAMS and EMSes have been around for a long time, with their increased penetration, the amount of generated data is triggering the adoption on big data and machine learning techniques that are integrated in applications serving all segments of the grid~\citep{wang2018review,8937810}. Data driven machine learning (ML) models are different than traditional statistical models in that they are able to \textit{automatically} learn an underlying distribution. However, to achieve that, a well defined knowledge discovery process (KDP) needs to be followed~\citep{fayyad1996knowledge}. The main steps of KDP consist of 1) data analysis, 2) data preparation (pre-processing), 3) model training and evaluation and 4) model deployment~\citep{ruf2021demystifying} as also represented in Figure~\ref{fig:data-infrastructure}. In the past, such process and the enabling tools were familiar only to a limited number of domain experts and the process involved intense manual effort. However, in the last five years, coordinated efforts have been taken by the private and public sectors to democratize AI and model development~\citep{allen2019democratizing} to empower less specialized users. 
	
	The democratization process involves a division of labour and automation like approach applied to the KDP, as elaborated more in details in~\citep{ruf2021demystifying}, where rather than a domain expert executing the step by step the process in Figure~\ref{fig:data-infrastructure} from start to end, they only need to control the process at a few key steps. For instance, to develop a home energy consumption prediction model, the users need carefully select the relevant data, also referred to as \textit{data model}, engineer the desired features, and configure the desired pipeline by selecting the ML methods to be applied and then selecting the best model to be deployed to production. Such automation is enabled by machine learning operations (MLOps)~\citep{ruf2021demystifying} and is being piloted in projects such as I-NERGY\footnote{\url{https://i-nergy.eu}} and MATRYCS\footnote{\url{https://matrycs.eu}}~\citep{en14154624}. 
	
	More recently, the authors in~\citep{patel2020unification} proposed a ``\textit{unification of machine learning features}'' in which a common/unified data preparation phase, that is the most time consuming and has a very large influence on the final ML model performance, is best automated by feature stores (see phase 2 in Figure~\ref{fig:data-infrastructure}). Such approach further reduces the time spent on the most time-consuming phase of ML application development, benefiting data scientists, engineers, and stakeholders. 
	While most of the MLOps automation steps are generic across domains, the ones concerned with data analysis and data preparation (pre-processing) have domain specifics and may significantly impact the model fairness and performance~\citep{cerar2021learning}. Assume aggregated home consumption is ingested from a smart meter and sent as is to train a ML algorithm. In such case, the model will learn the likely distribution of the values in the metering data and predict future energy consumption based on that, similar to the work in~\citep{kong2017short}. However, if additional weather data would also be used for training, the model would learn to associated lower energy consumption with sunny days and high temperatures, thus yielding superior performance. It is common practice in the literature to use such additional data. For instance, in \citep{zhang2018forecasting} they estimated consumption using timestamp (month, hour, week of year, day of week, season), energy consumption, weather (condition, severity, temperature, humidity), energy price. In \citep{Lim2020DeepLearning-BasedAnalysis}, the ML based estimation was done using timestamp, electricity contract type, energy consumption and city area. While taxonomies and semantic modelling efforts for representing various domains \citep{bertoli2022semantic}, including technical aspects of energy infrastructure  \citep{daniele2015created}, have been developed, a study that considers how to design data models for developing ML-based energy applications and the subsequent management of such features in development and production pipelines is missing.
	
	In this paper, we perform a first time study on data models and feature management solutions for developing ML-based energy applications. Prior work focused on structuring energy-only data for interoperability purposes resulting in ontologies such as SAREF~\citep{daniele2015created} and SARGON~\citep{haghgoo2020sargon} without considering a broader overview for supporting generic energy data modelling and management of features. The contributions of this paper are:
	\begin{itemize}
		\item A taxonomy for guiding the design of data models suitable for ML-based energy applications and identification of relevant data sources for the categories in the taxonomy. Using a consumption forecasting example, we show that 1) contextual features are found to be almost as important and domain specific features and 2) by adding additional contextual and behavioral aspects to the typical feature set decreases the prediction error (i.e. mean average error) by 11\% from 0.308 kWh to 0.274 kWh.
		\item Analysis feature management process using emerging feature store solutions. We show that compared to other solutions, the Feast feature store can take by up to 99 percentage points less time to process, enrich and obtain the features needed for production ready model development.
	\end{itemize}
	
	The rest of the paper is structured as follows. Section~\ref{sec:taxonomy} elaborates on the proposed taxonomy for designing energy data models,  Section~\ref{sec:feature-store} discusses feature management using feature stores while Section~\ref{sec:evaluation} details the evaluation and benchmarks. Finally, Section~\ref{sec:conclusions} concludes the paper.
	
	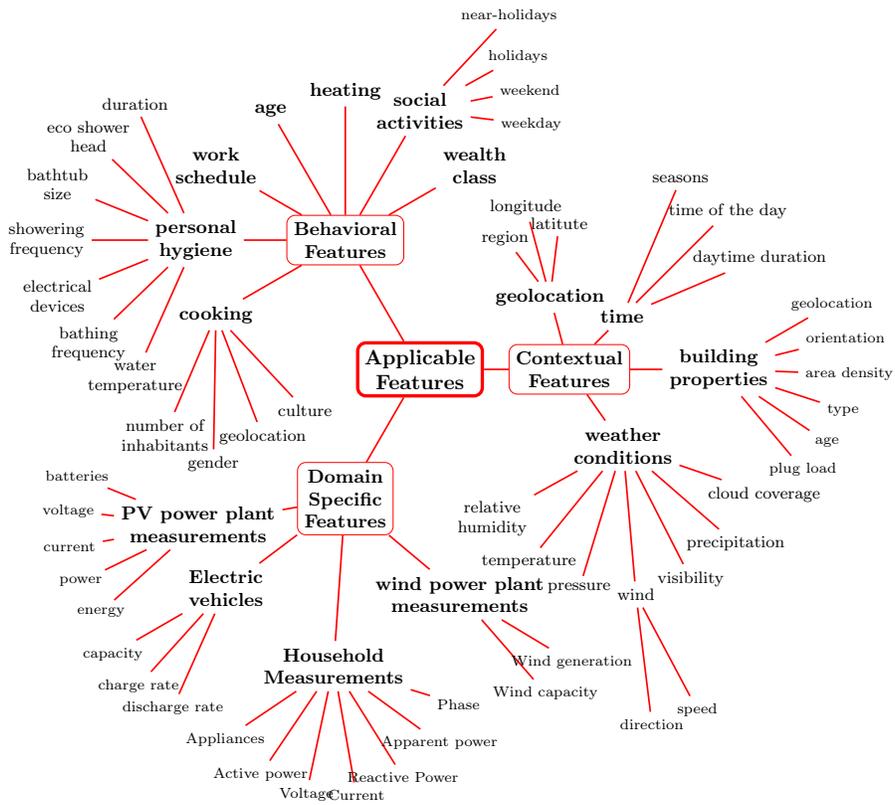
\begin{figure}[htbp]
		\centering
		\resizebox{\linewidth}{!}{
			\begin{tikzpicture}[
			scale=0.9,
			grow cyclic,
			edge from parent/.style = {draw, -, thick, red},
			%every node/.style = {font=\footnotesize},
			sloped,
			box/.style = { shape=rectangle, rounded corners, draw=red },
			level/.style = { align=center },
			level 0/.style = { level, font=\bfseries\normalsize, align=center, ultra thick },
			level 1/.style = { level, level distance=8em, font=\bfseries\small, sibling angle=120 },
			level 2/.style = { level, level distance=8em, font=\bfseries\small, sibling angle=30 },
			level 3/.style = { level, level distance=8em, font=\footnotesize, sibling angle=22 },
			level 4/.style = { level, level distance=7em, font=\scriptsize, sibling angle=18 }
			]
			\node [level 0, box] {Applicable\\Features}
			child [level 1] { node [box] {Domain\\Specific\\Features}
				child [level 2, sibling angle=40, rotate=10] { node {PV power plant\\measurements}
					child [level 4, sibling angle=16] { node{batteries} }
					child [level 4, sibling angle=16] { node{voltage} }
					child [level 4, sibling angle=16] { node{current} }
					child [level 4, sibling angle=16] { node{power} }
					child [level 4, sibling angle=16] { node{energy} }
				}
				child [level 2, sibling angle=40, rotate=-3] { node {Electric\\vehicles}
					child [level 4, rotate=11, sibling angle=18] { node{capacity} }
					child [level 4, rotate=11, sibling angle=18] { node{charge rate} }
					child [level 4, rotate=11, sibling angle=18] { node{discharge rate} }
				}
				child [level 2, sibling angle=40, rotate=6, level distance=9em] { node {Household\\Measurements}
					child [level 4, rotate=14, sibling angle=22] { node{Appliances} }
					child [level 4, rotate=14, sibling angle=22] { node{Active power} }
					child [level 4, rotate=14, sibling angle=22] { node{Voltage} }
					child [level 4, rotate=14, sibling angle=22] { node{Current} }
					child [level 4, rotate=14, sibling angle=22] { node{Reactive Power} }
					child [level 4, rotate=14, sibling angle=22] { node{Apparent power} }
					child [level 4, rotate=11, sibling angle=22] { node{Phase} }
				}
				child [level 2,sibling angle=40, rotate=20] { node {wind power plant\\measurements}
					child [level 4] { node {Wind capacity}}
					child [level 4] { node {Wind generation}}
				}
			}
			child [level 1] { node [box] {Contextual\\Features}
				child [level 2, level distance=5em, rotate=-10] { node {weather\\conditions}
					child [level 3, rotate=-30] { node {relative\\humidity}}
					child [level 3, rotate=-30] { node {temperature}}
					child [level 3, rotate=-30] { node {pressure}}
					child [level 3, rotate=-30] { node {wind}
						child [level 4, sibling angle=16, rotate=10] { node {direction}}
						child [level 4, sibling angle=16, rotate=15] { node {speed}}
					}
					child [level 3, rotate=-30] { node {visibility}}
					child [level 3, rotate=-30] { node {precipitation}}
					child [level 3, rotate=-30, level distance=8em] { node {cloud coverage}}
				}
				child [level 2, rotate=15] { node {building\\properties}
					child [level 4, sibling angle=16, rotate=-10] { node {plug load}}
					child [level 4, sibling angle=16, rotate=-10] { node {age}}
					child [level 4, sibling angle=16, rotate=-10] { node {type}}
					child [level 4, sibling angle=16, rotate=-10] { node {area density}}
					child [level 4, sibling angle=16, rotate=-10] { node {orientation}}
					child [level 4, sibling angle=16, rotate=-10] { node {geolocation}}
				}
				child [level 2, rotate=30, level distance=4em] { node {time}
					child [level 3] { node {daytime duration}}
					child [level 3] { node {time of the day}}
					child [level 3] { node {seasons}}
				}
				child [level 2, rotate=60, level distance=4em] { node {geolocation}
					child [level 3, rotate=0, level distance=4em] { node {latitute}}
					child [level 3, rotate=0, level distance=5em] { node {longitude}}
					child [level 3, rotate=0, level distance=4em] { node {region}}
				}
			}
			child [level 1] { node [box] {Behavioral\\Features}
				child [level 2] { node {wealth\\class}}
				child [level 2] { node {social\\activities}
					child [level 4, level distance=6em, rotate=-40] { node {weekday}}
					child [level 4, level distance=6em, rotate=-40] { node {weekend}}
					child [level 4, level distance=6em, rotate=-40] { node {holidays}}
					child [level 4, level distance=7em, rotate=-40] { node {near-holidays}}
				}
				child [level 2] { node {heating}}
				child [level 2] { node {age}}
				child [level 2] { node {work\\schedule}}
				child [level 2] { node {personal\\hygiene}
					child [level 3] { node {duration}}
					child [level 3] { node {eco shower\\head}}
					child [level 3] { node {bathtub\\size}}
					child [level 3] { node {showering\\frequency}}
					child [level 3] { node {electrical\\devices}}
					child [level 3] { node {bathing\\frequency}}
					child [level 3] { node {water\\temperature}}
				}
				child [level 2] { node {cooking}
					child [level 3, rotate=70, level distance=7em] { node {number of\\inhabitants}}
					child [level 3, rotate=70, level distance=8em] { node {gender}}
					child [level 3, rotate=70, level distance=7em] { node {geolocation}}
					child [level 3, rotate=70, level distance=7em] { node {culture}}
				}
			}
			;
			\end{tikzpicture}
		}
		\caption{A taxonomy of features relevant for energy application data model design.}
		\label{fig:ApplicableFeatures}
	\end{figure}

	\section{Taxonomy for Guiding Data Model Design}
	\label{sec:taxonomy}
	In this section, we propose a taxonomy that identifies and structures various types of data related to energy applications. Based on this taxonomy, data models can be designed and implemented in database-like systems or feature storing systems for ML model training. The proposed taxonomy is depicted in Figure~\ref{fig:ApplicableFeatures} and distinguishes three large categories: domain specific, contextual and behavioural. To achieve inter-database or inter-feature store interoperability, the proposed model can be encoded in semantically inter-operable formats using already available vocabularies and ontologies such as SAREF, SARGON and Schema.org for the domain specific features,  OWL, FOAF and other semantic structures available to create linked open datasets\footnote{Linked Open Data Vocabularies, https://lov.linkeddata.es/dataset/lov/vocabs} ontologies for behavioral and contextual features.  
	
	\subsection{Domain specific}
	Domain-specific features are measurements of energy consumption and production collected by IEDs installed at the various points of the energy grid. Additionally, information associated with energy-related appliances, such as the type of heating (\ie, heat pump, gas furnace, electric fireplace), can be presented in meta-data. In Figure~\ref{fig:ApplicableFeatures} we identify PV power plant generation, electric vehicles (EVs), wind power generation, and household consumption. Power plants data include battery/super-capacitor capacity, voltage, current, power, and energy measurements and can be found in at least two publicly available datasets as listed in Table~\ref{tab:datasets}. These datasets may also include metadata such as plant id, source key, geographical location, and measurements that fall under the contextual features group, such as air temperature. Wind power generation datasets contain the generated power and sometimes electrical capacity.
	
	With a recent spike in the popularity of EVs, power grids need to be designed/improved accordingly because of their significant energy capacity and power draw. In the third column of Table~\ref{tab:datasets}, we listed good quality publicly accessible datasets related to EVs. The datasets consider, for instance, battery capacity, charge rate, and discharge rate. 
	
	For household measurements, consumption may be measured by a single or several smart meters, thus being aggregated or per (set of) appliances. Depending on the dataset, they contain active, reactive, and apparent power, current, phase, and voltage. In some cases, meta-data about the geolocation, orientation, or size of the house may also be present. There are a number of good quality publicly accessible datasets, as can be seen from the fourth column of Table~\ref{tab:datasets}.
	
	\begin{table}[htbp]
		\ra{1.2}
		\caption{Datasets and studies suitable for the categories in the feature taxonomy.}
		\label{tab:datasets}
		\centering
		\begin{tabular}{|ccccc|}\hline
			\multicolumn{5}{|c|}{Domain specific features}\\
			PVs & EVs & Wind & \multicolumn{2}{c|}{Household}\\\hline
			
			\makecell[tl]{
				UKPN~\citep{UKPN}\\
				Kannal~\citep{KannalIN}
			}
			&
			\makecell[tl]{
				V2G~\citep{soares2013v2g}\\
				emobpy~\citep{morales2021emobpy}
			}
			&
			\makecell[tl]{
				Sandoval~\citep{SandovalGER}\\
				Lafaz~\citep{LafazGER}
			}
			&
			\makecell[tl]{
				BLOND-50~\citep{BLOND}\\
				FIRED~\citep{FIRED}\\
				UK-DALE~\citep{UKDALE}\\
				REFIT~\citep{REFIT}\\
				ECO~\citep{ECO}\\
				REDD~\citep{REDD}
			}
			&
			\makecell[tl]{
				iAWE~\citep{iAWE}\\
				COMBED~\citep{COMBED}\\ % !
				HUE~\citep{makonin2018hue}\\
				HVAC~\citep{grigore2019hvac}\\
				Synthetic~\citep{li2021synthetic}
			}
			\\\hline
		\end{tabular}
		\\[0.25em]
		\begin{tabular}{cc}
			\begin{tabular}[t]{|ccc|}\hline
				\multicolumn{3}{|c|}{Contextual features}\\
				\multicolumn{2}{|c}{Weather} & Building \\\hline
				
				\makecell[tl]{
					UKPN~\citep{UKPN}\\
					Kannal~\citep{KannalIN}\\
					Sandoval~\citep{SandovalGER}\\
					Lafaz~\citep{LafazGER}
				}
				&
				\makecell[tl]{
					ECN~\citep{UK-ECN}\\
					MIDAS~\citep{MIDAS}\\
					Keller\citep{KellerGER}\\
					Kukreja~\citep{KukrejaIN}
				}
				&
				\makecell[tl]{
					HUE~\citep{makonin2018hue}
				}
				\\\hline
			\end{tabular}
			&
			\begin{tabular}[t]{|c|}\hline
				\multicolumn{1}{|c|}{Behavioral features}\\
				\\\hline
				\makecell[lt]{
					Social~\citep{Spichakova2019FeatureEngineeringForShort-TermForecastOfEnergyConsumption}\\
					Wealth~\citep{Adika2012NorthAmericanPowerSymposium(NAPS)}\\
					Gender~\citep{parkinson2021overcooling}
				}
				\\\hline
			\end{tabular}
		\end{tabular}
	\end{table}
	
	\subsection{Contextual features}
	We refer to contextual features as measured data that is not directly collected by measuring device(s), however such data~\citep{Sinimaa2021FeatureEngineeringOfWeatherDataForShort-TermEnergyConsumptionForecast} may be critical in developing better energy consumption or production estimates. In Figure~\ref{fig:ApplicableFeatures} we identify 1) weather data such as wind speed/direction, temperature, relative humidity, pressure, cloud coverage and visibility, 2) building properties such as type of insulation, year of building, type of property, orientation, area density and 3) time related features such as part of the day and daytime duration. 
	
	Weather datasets, such as the ones listed in the fifth column of Table~\ref{tab:datasets}, usually provide numbers related to temperature and precipitation. Some also provide more climate elements like fog and hail, wind speed, humidity, pressure and sunshine/solar data. They also all provide the geographical location and the time period of measurement.
	\subsection{Behavioral features}
	Behavioral features refer to aspects related to people's behavior such as work schedule, age, wealth class and hygiene habits as per Figure~\ref{fig:ApplicableFeatures}. Several studies, such as ~\citep{Spichakova2019FeatureEngineeringForShort-TermForecastOfEnergyConsumption, Adika2012NorthAmericanPowerSymposium(NAPS), parkinson2021overcooling} show that electricity consumption is influenced by the behavior of the inhabitants. These are, for instance, personal hygiene, person's age, person's origins, work habits, cooking habits, social activities~\citep{Spichakova2019FeatureEngineeringForShort-TermForecastOfEnergyConsumption}, and wealth class~\citep{Adika2012NorthAmericanPowerSymposium(NAPS)}. People of different ages have diverse electricity consumption patterns because of the difference in sleep habits and lifestyles in general. Social activities such as holidays, near-holidays, weekdays, and weekends also make a difference in the electricity consumption in homes, offices, and other buildings. Wealth class also influences electricity consumption because it impacts the lifestyle. If we are talking about a home, personal hygiene also has a major impact. Showering or bathing can take a considerable amount of warm water, depending on the showering frequency, shower duration, and how much water the shower head pours each minute. Besides that, we have to consider electrical devices such as a fan, hairdryer, or infrared heater.
	
	Gender of the inhabitants also plays a role as men and women have different preferences when it comes to air temperature, water temperature~\citep{parkinson2021overcooling} and daily routines. While such data is useful for estimating and optimizing energy consumption, it raises ethical and privacy concerns. To some extent, such data can be collected through questionnaires, studies, or simulations to describe the behavioral specifics of users, groups, or communities.

	\section{Feature management for ML model development}
	\label{sec:feature-store}
	Traditional feature engineering for ML model development is often considered an art, requiring intense effort from expert data scientists and domain experts. Such teams develop training data containing raw or processed domain specific features with contextual features such as weather. One example row in such training data, also referred as feature vector, may look like: [\textit{timestamp, consumption[kWh], temperature, solar radiation}] where the consumption and weather metrics may represent instant values of available or averages over an hours, a day, etc. As averaging, especially over longer windows may remove some spikes that are relevant for accurate predictions such as consumption forecasting, additional statistics can be added to the training data~\citep{Ouyang2017}. For instance, adding max, min and standard deviation [\textit{timestamp, avg. consumption [kWh], min. consumption [kWh], max. consumption [kWh], std. dev. of consumption, temperature, solar radiation}] may lead to significant performance increases. The ML community developed tools\footnote{Time-series feature generator, https://tsfresh.readthedocs.io/en/latest/} that simplify the computation and development of such feature vectors that were traditionally saved to files on disk or tabular databases. Studies such as \citep{cerar2021learning} and \citep{Bertalanicimage} have shown that feature engineering through dimensionality reduction using statistical techniques such as discussed in this example and dimensionality expansion techniques can significantly increase the performance of the final model. 
	
	\begin{lstlisting}[language=Python, caption=Example recipe for generating the average household energy consumption over 1 hour feature (energy\_mean).]
	residential_hourly_stats = FileSource(
	path=str(residential_dataset_path),
	event_timestamp_column='timestamp',
	)
	
	consumption_hourly_stats_view = FeatureView(
	name='residential_hourly_stats',
	entities=['residential_id'],
	# Measurement validity period. 
	# 1h in our case. Used when data is joined
	ttl=Duration(seconds=3600),
	# Describe features in the data
	features=[
	Feature(name='ts', dtype=ValueType.FLOAT),
	Feature(name='energy', dtype=ValueType.FLOAT),
	],
	online=True,
	# source of the data (parquet file in our case)
	batch_source=residential_hourly_stats,
	tags={},
	)
	\end{lstlisting}
	
	\begin{figure}[htbp]
		\centering
		\includegraphics[width=0.9\linewidth]{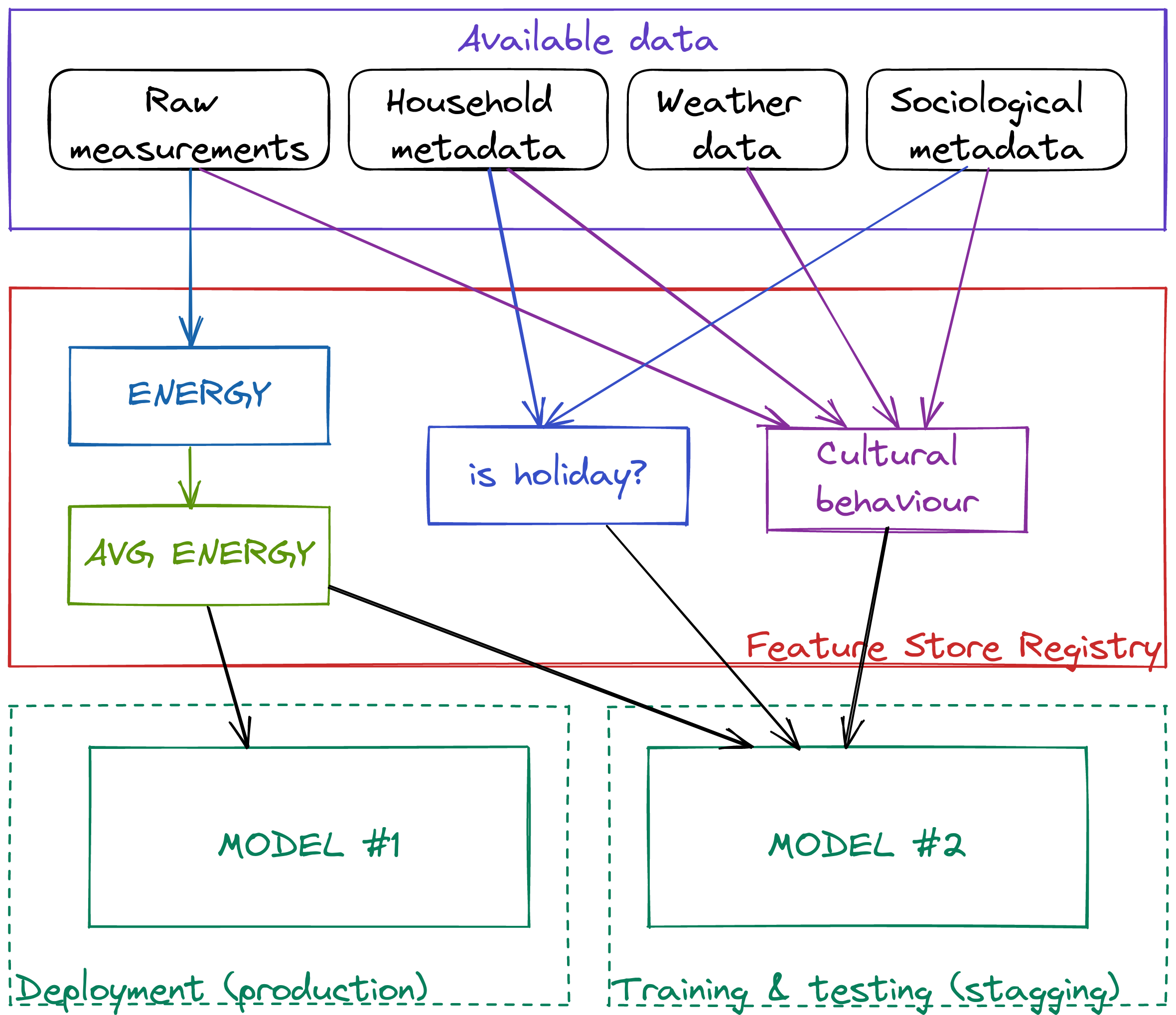}
		\caption{The ``data flow'' of transformation(s) in feature store. Model(s) in staging can be developed in parallel with a deployed model(s).}
		\label{fig:feature-store-concept}
	\end{figure}
	
	\begin{table*}[htbp]
		\caption{List of open-source feature store solutions suitable for time series.}
		\label{tab:available-feature-stores}
		\ra{1.3}
		\centering
		\footnotesize
		\begin{tabularx}{\linewidth}{@{}L{0.7}C{0.4}L{1.3}L{1.3}L{1.3}L{}L{}}
			\toprule
			\thead{Name}
			& \thead{Open\\Source}
			& \thead{Data Sources}
			& \thead{Offline Storage}
			& \thead{Online Storage}
			& \thead{Deployment}
			
			\\\midrule
			
			Feast
			& Y
			& BigQuery, Hive, Kafka, Parquet, Postgres, Redshift, Snowflake, Spark, Synapse
			& BigQuery, Hive, Pandas, Postgres, Redshift, Snowflake, Spark, Synapse, Trino, custom
			& DynamoDB, Datastore, Redis, Azure Cache for Redis, Postgres, SQLite, custom
			& AWS Lambda, Kubernetes, local
			\\\midrule
			
			Hopsworks
			& Y
			& Flink, Spark, custom Python, Java, or Scala connectors
			& Azure Data Lake Storage, HopsFS, any SQL with JDBC, Redshift, S3, Snowflake
			& any SQL with JDBC, Snowflake
			& AWS, Azure, Google Cloud, local
			\\\midrule
			
			Butterfree
			& Y
			& Kafka, S3, Spark
			& S3, Spark Metastore
			& Cassandra
			& local
			\\\midrule

		\end{tabularx}
	\end{table*}
	
	In emerging modern data infrastructures, the features produced using various dimensionality reduction and expansion techniques, to be used for model training, are managed by feature stores rather than manually managed through generation scripts and file/database storage. These are data management services that harmonize data processing steps producing features for different pipelines, making it more cost-effective and scalable compared to traditional approaches~\citep{patel2020unification}.
	
	A feature store's registry contains instructions on how every desired feature should be produced such as depicted in Listing 1 for hourly mean consumption. To develop such recipes, a taxonomy such as proposed in this paper is used to guide the data architect in the defining such features. Then corresponding data sources, such as identified in Table~\ref{tab:datasets}, should be used to build each desired feature according to the recipe. Once all the recipes are defined and all the data sources as available, the feature stores automatically manages the feature generation process as also depicted in Figure~\ref{fig:feature-store-concept}. When additional recipes and data sources are added to the store, they will be automatically generated and made available to training models. 
	
	As depicted in Figure~\ref{fig:data-infrastructure} and described in Section~\ref{sec:intro}, a feature store ingests data, transforms it according to the instructions kept in a feature store registry, and serves features. Within the feature store, as depicted in Figure~\ref{fig:feature-store-concept}, a flow starts where a feature store takes raw data discussed in Section~\ref{sec:taxonomy}. Inside the feature store, the data is transformed according to the instructions in a feature store registry. A feature can be built out of single or multiple sources or features. The flow ends with feature serving, where features are passed to the model(s), where each requires a different set of features to work correctly. Many different features can coexist simultaneously in a feature store, and a model may not use all features. This way, models can be developed using any combination of features and then the best model selected for deployment to production. The model development and selection is also automated using MLOps tools as discussed in Section \ref{sec:intro} and depicted in Figure~\ref{fig:data-infrastructure}.
	
	Several closed sources feature stores have been recently developed while to the best of our knowledge, only three that are suitable for time-series feature management are available as open source with their characteristics summarised in  Table~\ref{tab:available-feature-stores}. It can be seen from the third column of the table that they include connectors to support fast interconnection with various storage solutions (BigQuery, S3, Postgres...) and streaming platforms (Kafka, Spark). As per columns three and four, it can be seen that all open source stores support offline and online storage such as public cloud provider's BigQuery, Azure, S3 and Snowflake or open source solutions such as PostgreSQL and Cassandra.  As can be seen from the sixth column of the table, the open source feature stores can be deployed locally and also in the public cloud.
	
	\section{Evaluation}
	\label{sec:evaluation}
	% Data model and feature store benchmarks
	To illustrate the need for the proposed taxonomy in developing rich feature sets, their influence on model development we develop and evaluate the importance of an extensive list of features (i.e. data model). It should be noted that the feature importance in MLOps pipelines is assessed by the Model Training component depicted in Figure \ref{fig:data-infrastructure} subject to the availability of the features. 
	
	Besides the obvious feature management automation convenience introduced by feature stores, we also quantify the performance of Feast, one of the available open source solutions, against more traditional management solution in timings relevant to generating the features developed in this work. 
	
	\begin{figure}[htbp]
		\begin{center}
			\includegraphics[width=0.9\linewidth]{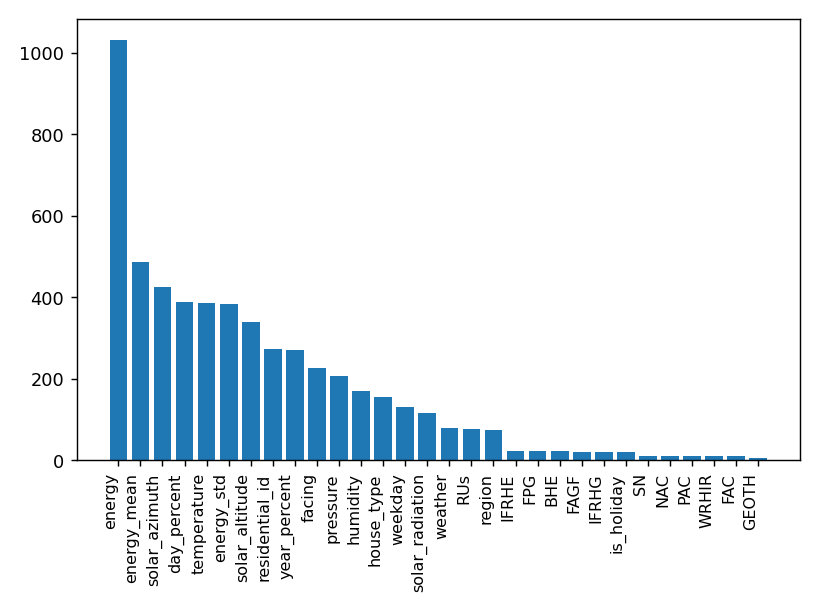}
		\end{center}
		\ra{0.75}
		\tiny\begin{tabular}{ll}
			\multicolumn{2}{c}{\bfseries Legend:}\\
			
			\multicolumn{2}{l}{\bfseries raw}\\
			energy & current hourly energy consumption\\
			
			\multicolumn{2}{l}{\bfseries statistical}\\
			energy\_mean & rolling hourly average; window size 10 hours\\
			energy\_std & standard deviation; window size 10 hours\\
			
			\multicolumn{2}{l}{\bfseries weather}\\
			temperature & current outdoor temperature\\
			humidity & current outdoor relative air humidity\\
			pressure & current outdoor air pressure\\
			weather & weather condition (\eg, cloudy, rain, snow)\\
			solar\_altitude & Sun altitude (VSOP~87 model)\\
			solar\_azimuth & Sun azimuth (VSOP~87 model)\\
			solar\_radiation & clear sky radiation (VSOP~87 model)\\
			
			\multicolumn{2}{l}{\bfseries building properties}\\
			residential\_id & residential unique ID\\
			house\_type & house type (\eg, apartment, duplex, bungalow)\\
			facing & house orientation (\eg, North, South, Northwest)\\
			RUs & number of rental units\\
			%EVs & total battery capacity of EV(s)\\
			SN & special operation regimes\\
			FAGF & forced air gas furnace\\
			%HP & heat pump (incl. a/c)\\
			FPG & gas fireplace\\
			%FPE & electric fireplace\\
			IFRHG & in-floor radiant heating (gas boiler)\\
			NAC & no a/c\\
			FAC & has fixed a/c unit\\
			PAC & has portable a/c unit\\
			BHE & baseboard heater (electric)\\
			IFRHE & in-floor radiant heating (electric)\\
			WRHIR & water radiant heat (cast iron radiators)\\
			GEOTH & geothermal\\
			
			\multicolumn{2}{l}{\bfseries time}\\
			day\_percent & percentage of the day elapsed\\
			year\_percent & percentage of the year elapsed\\

			\multicolumn{2}{l}{\bfseries sociological}\\
			is\_holiday & is a holiday\\
			weekday & integer presentation of day of week day\\

			region & geographic region\\
		\end{tabular}
		
		\caption{Feature importance score for estimating future (1h ahead) energy consumption.}
		\label{fig:reach:importance}
	\end{figure}
	
	Throughout the section we use HUE (the Hourly Usage of Energy dataset for buildings in British Columbia) dataset mentioned in Table \ref{tab:datasets}. It contains hourly data from 28 households in Canada, collected in different timespans between 2012 and 2020. The dataset consists of raw data, household metadata, and weather data ($\approx$744\,000 samples in total). This dataset is suitable for analyzing and predicting household energy consumption.
	
	\subsection{Feature importance}
	To asses the importance of the three categories of features from the proposed taxonomy in Figure~\ref{fig:ApplicableFeatures}, in addition to the HUE dataset that includes domain specific, contextual and behavioural features, we also consider additional contextual features related to solar radiation and altitude produced by a model~\citep{reda2004solar}.
	
	From HUE we have the energy consumption measured by IEDs and categorical variables related to the type of heating devices such as forced air gas furnace (FAFG), heat pump (HP), etc. as domain measurement. Additionally, as contextual features we consider available metadata related to the building such as the id of the residence, house orientation, type of house related to the geographical location such as region and meteorological data such as pressure, temperature, humidity and weather (e.g. cloudy, windy, snow storm). We also consider behavioural features related to weekdays and holidays (is\_holiday, weekday, is\_weekend). To understand the importance these features may have in estimating short time consumption for 1 hour ahead we train an XGBoost\footnote{\url{https://xgboost.readthedocs.io/en/stable/}} regressor and asses the assigned importance. 
	
	The results of the feature importance as learnt by XGBoost are presented in Figure~\ref{fig:reach:importance}. It can be seen that the raw instant energy consumption is the feature that contributes the most to the energy estimation. The second and sixth most important features are the mean and standard deviation of the energy generated using statistical feature engineering techniques. Contextual features such as solar azimuth and how much of the 24h in a day have passed (day\_percent) are the third and fourth most important features. It can be noticed that the XGBoost considers the instant energy consumption more than twice as important as its rolling window average with a score of 1035 compared to 488. The importance of the second and third features are between 400 and 500, the importance of fourth to seventh features is also comparable, with values between 300 and 400. Starting with the eight feature, the importance decreases more abruptly from just below 300 to below 100 while the last 12 features, mostly related to the type of heating and cooling devices used as can be seen from the legend of Figure~\ref{fig:reach:importance}, are relatively less important by an order of magnitude lower than the first. Other features, which were omitted from bar plot, show no significant importance.
	
	\begin{table}[htbp]
		\caption{Impact of additional feature categories on the XGBoost regression model.}
		\label{tab:benchmark:features}
		\ra{1.3}
		\centering
		\footnotesize
		\begin{tabular}{@{}lrrr@{}}\toprule
			\thead{Feature set}
			& \thead{MSE [kWh]}
			& \thead{MAE [kWh]}
			& \thead{medAE [kWh]}
			\\\midrule

			raw & 0.343 & 0.317 & 0.146 \\

			above + statistical & 0.327 & 0.308 & 0.148 \\

			above + weather & 0.293 & 0.292 & 0.148 \\

			above + building properties & 0.265 & 0.278 & 0.141 \\

			above + time & 0.260 & 0.275 & 0.138 \\

			above + geolocation & 0.259 & 0.274 & 0.138 \\

			above + sociological & 0.258 & 0.274 & 0.138 \\
			
			\bottomrule
		\end{tabular}
	\end{table}
	\subsection{Impact of features on the model performance.}
	This section examines how each category of features from the proposed taxonomy in Figure~\ref{fig:ApplicableFeatures} can contribute to the model's accuracy. The goal was an accurate prediction of energy consumption 1 hour ahead. The training data was shuffled, split using 10-fold cross-validation, and evaluated 10-times using the XGBoost regressor algorithm. Every step of the ML pipeline was seeded for a fair comparison.
	
	Table~\ref{tab:benchmark:features} presents the evaluation on the impact of features on model performance. From top to bottom, each row adds a set of features. The "raw" feature set contains only instant energy consumption collected by IEDs. The "statistical" feature set adds rolling average and standard deviation for the last 10 hours. The weather feature set adds attributes regarding outside temperature, humidity, pressure, weather condition, theoretical solar altitude, azimuth and radiation. The building properties adds attributes of each household, house type, house facing direction, number of EVs, and type of heating system. The "time" feature set adds the percentage of day, week, and year elapsed. The "geolocation" set adds geographical longitude and latitude. Finally, the "sociological" feature set adds information regarding holidays, weekday, weekend, and information about region.
	
	Table~\ref{tab:benchmark:features} shows that adding additional features to the data significantly improve the estimation performance. Improvement can be observed through a consistent decrease of mean squared error of prediction (top to bottom).
	
	The first row shows that using (only current) raw values, the model achieves 0.343\,kWh mean squared error. By adding the "stastistical" feature set, MSE decreases to 0.327\,kWh. The most significant improvement is observed when weather data is added, where MSE drops from 0.327 to 0.292\,kWh. By adding building properties in addition to raw, statistical, and weather data, MSE further decrease to 0.265\,kWh. By adding time feature set, MSE decrease to 0.260\,kWh. Finally, a minor improvement comes from geolocation and sociological feature sets, where MSE decrease from 0.260 to 0.258\,kWh.
	
	\subsection{Benchmark in feature management solutions}
\begin{figure}[htbp]
		\centering
		\includegraphics[width=0.9\linewidth]{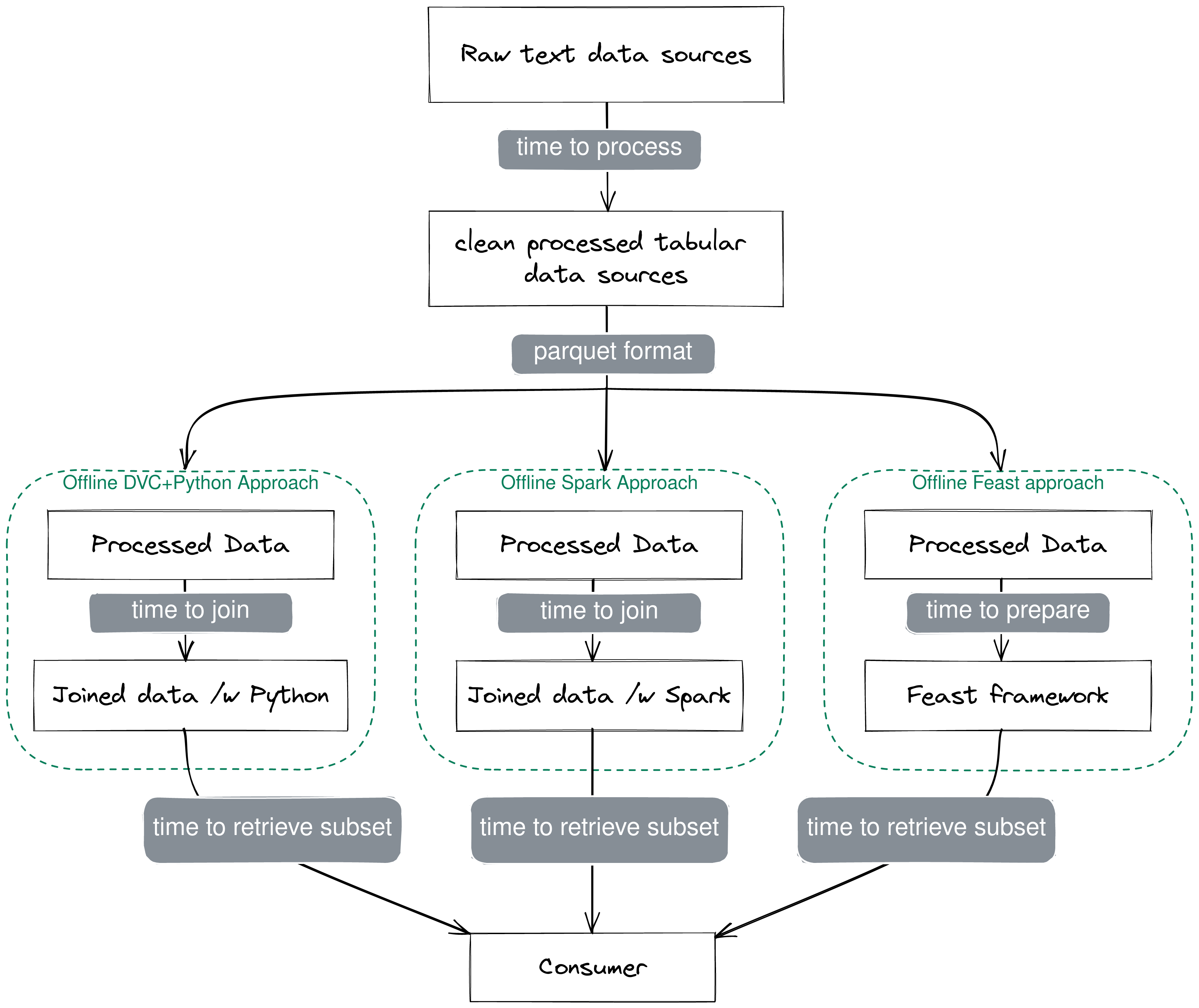}
		\caption{Benchmarking feature management solutions.}
		% https://www.sciencedirect.com/science/article/pii/S2212017312002836
		\label{fig:pipelines}
	\end{figure}

	To manage the features, we consider three approaches and evaluate their relative performance with respect to basic steps in the feature processing pipeline as illustrated in Figure~\ref{fig:pipelines}. As presented on the top of Figure~\ref{fig:pipelines}, all three pipelines share a common pre-processing step, where python scripts perform ``basic'' data cleaning of raw measurements and metadata and store data in structured Apache Parquet format. Parquet is a common data source format for feature stores as can also be seen from Table~\ref{tab:available-feature-stores}.
	
	In the first benchmarked feature management solution in Figure~\ref{fig:pipelines}, we used Python with Pandas and Data Version Control (DVC) tools. Pandas is a specialized tool for manipulating tabular data, and the DVC tool helps define and execute reproducible multi-step pipelines. The second benchmarked management solution uses Apache Spark controlled by pySpark. Spark runs on a single node, where Master and Worker nodes reside in their own Docker containers, which are interconnected with a subnetwork. Finally, the third benchmarked feature management solution is the Feast feature store that is used to access features. Feast accesses cold storage (i.e. Parquet files) on each request to deliver requested features.

	\begin{table}[htbp]
		\centering
		\footnotesize
		\begin{threeparttable}
			\caption{The benchmark results of the three pipelines on the HUE dataset.}
			\label{tab:implementation:evaluation}
			\begin{tabular}{@{}lrrr@{}}
				\toprule
				\multirow{2}{*}{\thead{Implementation}}
				& \multicolumn{3}{c}{Timings}\\\cline{2-4}
				& \thead{to process}
				& \thead{to join \& enrich}
				& \thead{to obtain subset}
				\\\midrule
				
				Python (Pandas)
				& $<$1\,s
				& 1707.308\,s
				& 0.617\,s
				\\
				
				Spark (pySpark)
				& -
				& 1050.897\,s
				& 6.941\,s
				\\
				
				Feature Store (Feast)
				& -
				& 1.235\,s
				& 20.581\,s
				\\\bottomrule
			\end{tabular}
			\begin{tablenotes}
				\footnotesize
				\item Note: \textit{Benchmarked on AMD Ryzen 5 3600 (6c/12t), 32GB DDR4 3200MHz, Samsung NVMe storage, Ubuntu Server 20.04}
			\end{tablenotes}
		\end{threeparttable}
	\end{table}
	
	% TODO (Gregor): Text is not up to date.
	
	The results of the benchmark  summarized in Table~\ref{tab:implementation:evaluation}. The common preprocessing step that deals with ``basic'' data cleaning of raw measurements and metadata, and stores data into structured Apache Parquet format takes less than a second. This relatively short execution time is due to the size of the data of up to 744\,000 rows. 
	
	The join\&enrich steps with Python took the longest to complete. It took Python 1707 seconds using Pandas to merge (\ie, SQL LEFT JOIN operation) three tables together and generate new features. However, pure Python was the fastest at retrieving a subset of the merged dataset, which took 0.6 seconds. For cases when an intermediate Parquet file with all the features exceeds the system memory size, it requires extra engineering and may not scale well.
	
	The approach with Spark was the fastest at merging datasets taking 1051 seconds, which is approximately 3 minutes faster than the pure Python approach. Faster execution is because most of the tabular data operations on Spark can utilize multiple threads and multiple workers (distributed). However, the distributed approach comes with an overhead of synchronization between workers and controller nodes, especially when flushing the output to the storage. Because of this overhead, Spark took longer, 7 seconds, to retrieve the subset of data. However, Spark would scale better with a large intermediate dataset. This is because Spark can scale in the number of workers and utilize distributed filesystems, such as HDFS and GlusterFS.
	
	The approach with feature store is a bit different from the pure Python and Spark approaches. It took Feast to "merge" the datasets at around one second. However, Feast does not "merge" anything at the preparation phase. Instead, it checks intermediate files (\ie, Parquet files from the first stage) and constructs data samples only when requested at the retrieval stage. The burden of joining data is pushed to the retrieval phase, which is why Feast requires the longest to retrieve the subset at approximately 21 seconds. While this is the slowest retrieval time, adding hot storage (\eg Redis) can be significantly improved, and it is expected in official documentation to be used in production deployment.
	
	One significant benefit of feature store (\ie, Feast) is handling new incoming data. Feature store would require only processed files to be updated before new data can be accessed. The present Python and Spark approach would have to redo the merging of the intermediate dataset with all features before new samples are accessible. 
	
	\section{Conclusions}
	\label{sec:conclusions}
	In this paper we proposed an taxonomy developed to empower designing data models for emerging ML based energy applications. Unlike prior works that focused on structuring energy data in various semantic formats for interoperability purposes, this work aims at data model development and subsequent feature engineering. The three main categories of features identified in the taxonomy are: behavioral, contextual and domain specific. We then discuss feature management solutions with focus on emerging feature stores that represent an important development in MLOps used in production systems.  Using a consumption forecasting example, we show that 1) contextual features are found to be almost
	as important and domain specific features and 2) by
	adding additional contextual and behavioral aspects to
	the typical feature set decreases the prediction error (i.e.
	mean average error) by 11\% from 0.308 kWh to 0.274
	kWh.
	
	We also prototyped and evaluated three complementary feature management solutions and showed that an open-source feature store solution can significantly reduce the time needed to develop new data models. Compared to currently used solutions, feature store can take by up to 99 percentage points less time to process, enrich and obtain the features needed for production ready model development.

	\section*{Acknowledgment}
	This work was funded by the Slovenian Research Agency under the Grant P2-0016 and the European Commission under grant number 872613.
	
	\bibliographystyle{elsarticle-num}
	\bibliography{biblio}

\end{document}